\documentclass{article}
\pdfoutput=1

\usepackage[preprint, nonatbib]{neurips_2022}

\usepackage[utf8]{inputenc}     
\usepackage[T1]{fontenc}        
\usepackage{hyperref}           
\usepackage{url}                
\usepackage{booktabs}           
\usepackage{amsfonts}           
\usepackage{amsmath}
\usepackage{nicefrac}           
\usepackage{microtype}          
\usepackage[dvipsnames, table]{xcolor}
\usepackage{authblk}
\usepackage{graphicx}           
\usepackage{array}              
\usepackage{multirow}
\usepackage[ruled]{algorithm2e} 
\usepackage{amssymb}            
\usepackage[noend]{algpseudocode}

\newcolumntype{M}[1]{>{\centering\arraybackslash}m{#1}!{\vrule width -1pt}}
\definecolor{BaselineShade}{gray}{0.88} \definecolor{AFLShade}{gray}{0.96}

\title{Adversarial Focal Loss: \\Asking Your Discriminator for Hard Examples}

\author[ ]{\textbf{Chen~Liu}\thanks{Project lead. Correspondence to: \url{chen.liu.cl2482@yale.edu}.}}
\author[ ]{\textbf{Xiaomeng~Dong}}
\author[ ]{\textbf{Michael~Potter}}
\author[ ]{\textbf{Hsi-Ming~Chang}}
\author[ ]{\textbf{Ravi~Soni}}
\affil[ ]{\textbf{GE Healthcare}}
\affil[ ]{\url{{FirstName}.{LastName}@ge.com}}

\begin{document}

\maketitle

\begin{figure}[htb!]
  \centering
  \includegraphics[width = 396pt]{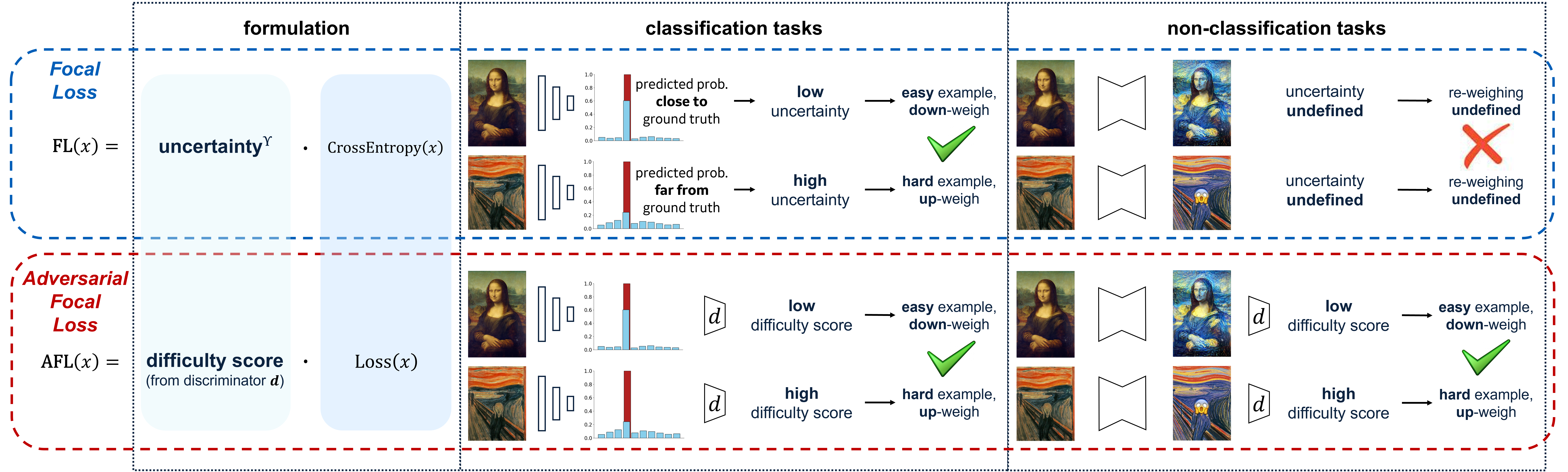}
  \caption{\textbf{Comparison between Focal Loss~(FL) and the proposed Adversarial Focal Loss~(AFL).} Both utilize a multiplicative coefficient to re-weigh examples based on their uncertainty/difficulty. However, AFL is more generalizable compared to Focal Loss. (1) AFL can be integrated into a variety of base loss functions (e.g., L1 loss, L2 loss, dice loss, keypoint accuracy loss, etc.) in addition to the Cross Entropy loss in the original formulation of Focal Loss. (2) Unlike Focal Loss, since AFL does not rely on the uncertainty from a classifier, it can be useful in tasks besides classification. Details about the formulations of FL and AFL can be found in section~\ref{sec_formulation_fl} and section~\ref{sec_formulation_afl}, respectively.}
  \label{fig_fl_vs_afl}
\end{figure}

\begin{abstract}
Focal Loss has reached incredible popularity as it uses a simple technique to identify and utilize hard examples to achieve better performance on classification. However, this method does not easily generalize outside of classification tasks, such as in keypoint detection. In this paper, we propose a novel adaptation of Focal Loss for keypoint detection tasks, called Adversarial Focal Loss~(AFL). AFL not only is semantically analogous to Focal loss, but also works as a plug-and-chug upgrade for arbitrary loss functions. While Focal Loss requires output from a classifier, AFL leverages a separate adversarial network to produce a difficulty score for each input. This difficulty score can then be used to dynamically prioritize learning on hard examples, even in absence of a classifier. In this work, we show AFL's effectiveness in enhancing existing methods in keypoint detection and verify its capability to re-weigh examples based on difficulty.

\end{abstract}

\section{Introduction}

Machine learning, similar to human learning, is trying to draw high-level knowledge from accessible data. Just as no two leaves are alike, the values of different data samples may also vary. Surprisingly, in machine learning it's not uncommon to turn a blind eye to such intrinsic differences in data, since it is often unclear how we shall treat data differently and doing so may be prohibitively labor intensive.

Arguably the most obvious form of the intrinsic disparity among data samples is the so-called \textit{class imbalance}~\cite{ClassImbalanceExp1, ClassImbalanceExp2, ClassImbalanceExp3} problem. It refers to the scenario where data samples of different classes are presented with significantly unequal frequencies. If the problem is left unattended, the resulting machine learning models can get heavily biased towards predicting majority classes.

Fortunately, this problem has attracted progressive research attention over the years. There are broadly two categories of solutions to this problem: data-level solutions and algorithm-level solutions. The former either down-samples the majority class or up-samples the minority classes through various sampling strategies~\cite{ClassImbalanceDataRand1, ClassImbalanceDataRand2, ClassImbalanceDataSoph1, ClassImbalanceDataSoph2}. The latter adjusts the training process or optimization objective~\cite{ClassImbalanceAlg1, ClassImbalanceAlg2, FocalLoss}. Among the algorithm-level solutions, a highly recognized method is Focal Loss~\cite{FocalLoss}.

Focal Loss is a loss function that dynamically scales a Cross Entropy objective based on the classifier's uncertainty for each sample. The uncertainty is measured by the difference between the predicted probability and the ground truth label. Focal Loss naturally solves the class imbalance problem by emphasizing minority-class samples that have higher uncertainties during optimization. In addition to solving class imbalance problems, the Focal Loss philosophy -- using uncertainty to re-weigh data samples -- can be used to tackle more generic data disparity challenges~\cite{FocalLossApp1, FocalLossApp2, FocalLossApp3, FocalLossApp4, FocalLossApp5, FocalLossApp6}.

Despite all its merits, Focal Loss is not well-defined outside classification tasks because it uses the classifier output as an uncertainty measure which is not readily available in most non-classification tasks. One such task is keypoint detection, a popular and challenging field in computer vision. Keypoint detection has diverse applications, such as facial expression recognition~\cite{KeypointFace1, KeypointFace2}, human pose estimation~\cite{KeypointPose1, KeypointPose2, KeypointPose3}, and medical diagnosis on anatomical abnormalities~\cite{KeypointMedical1, KeypointMedical2}. Meanwhile, keypoint detection is known to suffer from class imbalance, scale imbalance, long-tailed data distributions, etc.~\cite{KeypointImbalance1, KeypointImbalance2, KeypointImbalance3}. Focal Loss would be a great solution to these problems, if only its utilities were not restricted to classification tasks.

To overcome this issue, we propose a simple yet effective solution named Adversarial Focal Loss~(AFL) for keypoint detection. Rather than relying on existing model outputs, we leverage an external adversarial network which can holistically assess the difficulties of each data sample regardless of base model architecture. This in turn allows us to formulate a loss function that is semantically analogous to Focal Loss while cleanly extending to the keypoint detection problem. From our experiments, AFL can be seamlessly integrated into arbitrary loss functions. It minimally increases computational cost during training, and it adds no computational overhead during inference.

\section{Related Work}
\paragraph{Keypoint Detection}
The term ``keypoint detection'' has two slightly different meanings in computer vision. In the first definition, keypoint detection identifies \textit{representative, yet not pre-defined} points in an image, usually over multiple frames of a video, for purposes such as registration or simultaneous localization and mapping~(SLAM). This is usually done by using unsupervised image-processing techniques called keypoint detectors, with some of the most famous ones being SIFT~\cite{SIFT}, SURF~\cite{SURF}, and ORB~\cite{ORB}. This definition is out of the scope of this paper.

In the second definition, which this paper focus on, keypoint detection identifies \textit{pre-defined} points in an image that usually have precise meanings -- for example, a set of recognizable points on a face or well-defined joints in a body. This is usually accomplished using supervised methods~\cite{Hourglass, FPN, MaskRCNN, CPN, OpenPose, AssociativeEmbedding, PersonLab, MultiPoseNet, PoseResNet, PoseHRNet}.

Keypoint detection algorithms must answer two questions:
\begin{enumerate}
    \item How to represent keypoint locations?
    \item How to associate keypoints with objects when multiple objects are present?
\end{enumerate}

In answer to the first question, while a few prior research directly performed regression on pixel coordinates~\cite{KeypointRepreOther1, KeypointRepreOther2}, the mainstream approach is to represent keypoints using feature maps (usually called ``heatmaps'')~\cite{Hourglass, FPN, MaskRCNN, CPN, OpenPose, AssociativeEmbedding, PersonLab, MultiPoseNet, PoseResNet, PoseHRNet}. In the feature-map approach, if we aim to predict a total of $K$ target keypoints, we will predict $K$ feature maps, one for each keypoint. Next, keypoint locations can be discovered through post-processing techniques, such as using the location of the highest-intensity pixel over each feature map. A keypoint is considered missing/non-existent if the corresponding feature map contains all zeros after thresholding.

In answer to the second question, keypoint detection approaches fall into two categories: top-down and bottom-up. Bottom-up approaches~\cite{OpenPose, AssociativeEmbedding, PersonLab, MultiPoseNet} first detect all keypoints without caring about which object they belong to and then group the keypoints into objects. Top-down approaches~\cite{MaskRCNN, CPN, PoseResNet, PoseHRNet} identify objects first and then detect keypoints for each object. We demonstrate AFL on several common top-down approaches, though it could also be applied to bottom-up approaches.

\paragraph{Focal Loss}
\label{sec_formulation_fl}
Focal Loss, when it was first introduced~\cite{FocalLoss}, was presented as an upgrade to the Cross Entropy loss function in classification problems. Since the Cross Entropy loss comes with symmetric components, the authors of the Focal Loss paper used a simplification of notation as follows:
\begin{equation}
    p_t = 
    \begin{cases}
      p & \text{if ground truth label } y \text{ = 1}\\
      1-p & \text{if ground truth label } y \text{ = 0}
    \end{cases}
    \label{eq_pt}
\end{equation}
where $y$ is the one-hot encoded ground truth label and $p$ is the predicted probability after the activation. It can be observed that $(1-p_t)$ is the gap between the predicted probability score and the binary ground truth label, and hence it represents the uncertainty of the classifier on the input sample. With this notation, the Cross Entropy loss and the Focal Loss can be respectively written as $\textrm{CE}$~(Eq.~\ref{eq_ce}) and $\textrm{FL}$~(Eq.~\ref{eq_fl}).

\begin{equation}
    \textrm{CE}(p_t) = -\log{(p_t)}
    \label{eq_ce}
\end{equation}

\begin{equation}
\begin{split}
    \textrm{FL}(p_t) &= -(1-p_t)^{\gamma}\log{(p_t)}\\
    &= (1-p_t)^{\gamma}\textrm{ CE}(p_t)
\end{split}
    \label{eq_fl}
\end{equation}

The key element that sets Focal Loss apart from Cross Entropy is the introduction of the multiplicative coefficient $(1-p_t)^{\gamma}$ which leverages the uncertainty $(1-p_t)$. Easy examples lead to low uncertainty~(i.e., when $p$ is close to the ground truth label, or equivalently, when $(1-p_t)$ is close to 0), whereas hard examples lead to high uncertainty. As a result, this coefficient selectively down-weighs the well-classified easy examples and up-weighs the high-uncertainty hard examples. $\gamma$ is a tunable parameters that adjusts the strength of re-weighing.

\paragraph{Adversarial Networks}
The notion of an adversarial network was first conceptualized in the ``Generative Adversarial Network''~(GAN) paper~\cite{GAN}. A GAN consists of two components competing against each other, a generator~$\pmb{g}$ and a discriminator~$\pmb{d}$ (also known as the adversarial network). In its original formulation, the objective is to train $\pmb{g}$ to generate realistic data, such as an image of a human face, from random noise. While a cohort of real data is 
available, there is no meaningful one-to-one mapping between the noise vectors and the real data. In absence of paired ground truth, the adversarial network $\pmb{d}$ acts as the loss function. Iteratively, $\pmb{d}$ is trained to distinguish between what $\pmb{g}$ generated and the real data, and $\pmb{g}$ is trained to generate data that look real enough to fool $\pmb{d}$. If a delicate balance is well maintained over time, $\pmb{g}$ can gradually generate data that are indistinguishable from the real distribution.

In this paper, different from the original purpose of GANs, we do not aim to perform data generation in any form. In our case, the generator $\pmb{g}$ is replaced by the keypoint detection network $\pmb{f}$, while the discriminator $\pmb{d}$ is used to calculate the Adversarial Focal Loss. Specifically, we ask $\pmb{d}$ to judge the quality of each predicted set of keypoints, which indirectly reflects the difficulty score of the corresponding sample. The difficulty score can then be used to create a loss function for keypoint detection that is semantically analogous to Focal Loss. 

\section{Adversarial Focal Loss}
\label{sec_formulation_afl}

\begin{figure}[b!]
  \centering
  \includegraphics[width = 320pt]{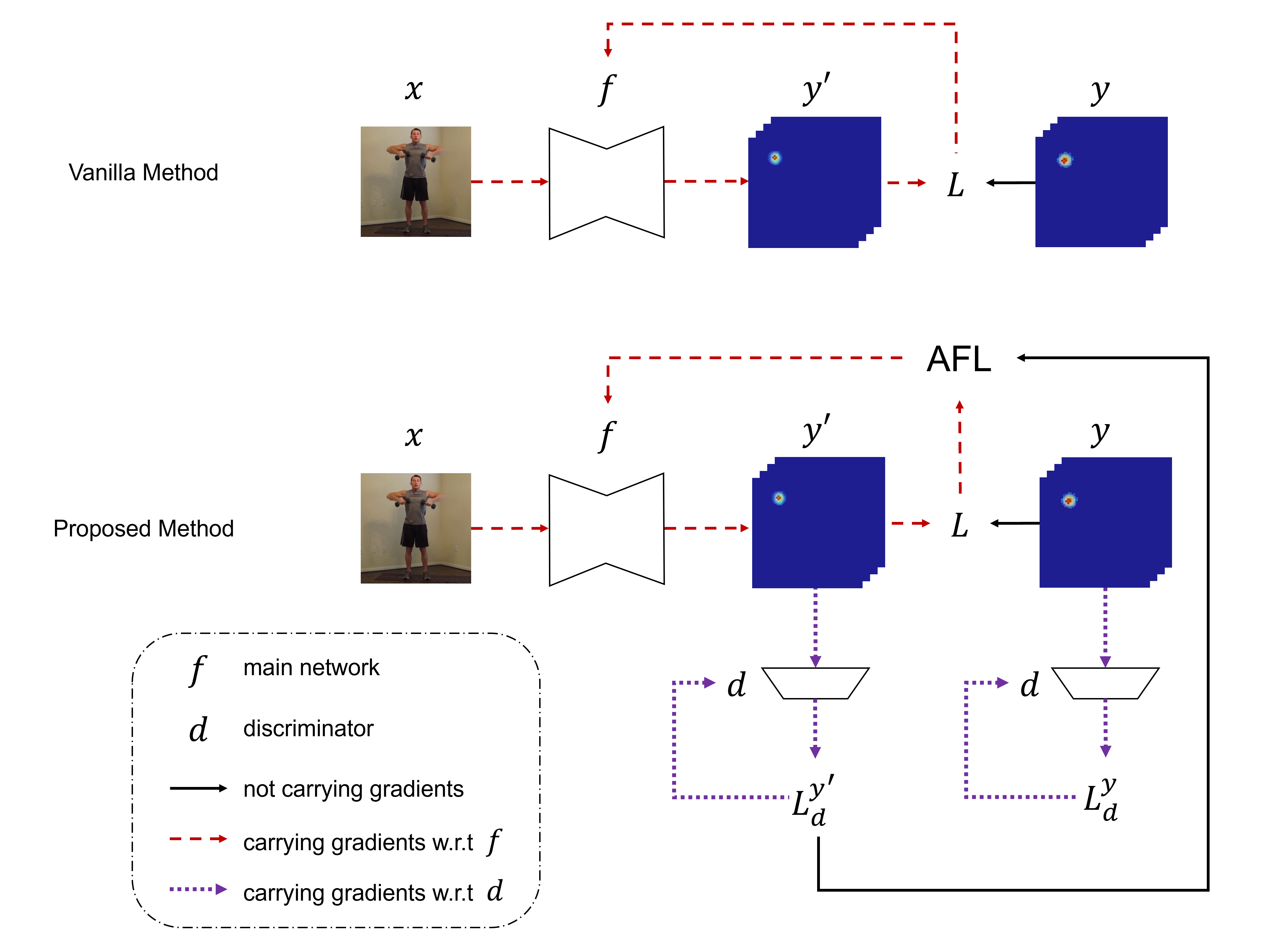}
  \caption{
  \textbf{Illustration of Adversarial Focal Loss (AFL).} AFL uses the discriminator ($\pmb{d}$) to produce a difficulty score~$\sigma(L_d^{y'})$ for each input sample. AFL subsequently utilizes the difficulty score to re-weigh the loss. This design is semantically analogous to Focal Loss even in absence of a classification objective.
  $L_d^{y'} = -\pmb{d}(y') = -\pmb{d}(\pmb{f}(x))$; $L_d^{y} = -\pmb{d}(y)$; $\sigma(\cdot)$: sigmoid function.
  In keypoint detection tasks only, feature representations $y'$ and $y$ are further condensed through a topology extractor (not shown in figure, see Appendix section~\ref{appx_topo_extractor}) before being given to $\pmb{d}$.
  }
  \label{fig_afl_framework_kp}
\end{figure}

Adversarial Focal Loss~(AFL) works in a similar way as the Focal Loss, except for one key difference: the way hard examples are discovered. Focal Loss uses the classification prediction as an indicator of uncertainty, whereas AFL leverages an independent discriminator to estimate the difficulty.

\paragraph{Vanilla Method for Keypoint Detection}
The workflow for vanilla keypoint detection is shown in Fig.~\ref{fig_afl_framework_kp} top. A main network $\pmb{f}$ is trained to predict feature maps $y'$ from the input image $x$. For each input $x$, a loss $L(x)=L(y, y')$ is computed between the prediction $y'$ and the ground truth $y$. The loss is then used to update $\pmb{f}$.

\paragraph{Adversarial Focal Loss}
The key of the proposed Adversarial Focal Loss method~(see Fig.~\ref{fig_afl_framework_kp}~bottom) is the discriminator~$\pmb{d}$. We use $\pmb{d}$ to judge the quality of a feature representation. $\pmb{d}$ is trained such that all model-generated representations are labeled as ``low-quality'' and all ground truth counterparts are labeled as ``high-quality''. Our formulation of the discriminator is inspired by Wasserstein GAN~(WGAN)~\cite{WGAN}, a widely-used adversarial network formulation that improved the stability of learning compared to the traditional GAN~\cite{GAN}.

Under the WGAN formulation~\cite{WGAN}, when a feature map $y'$ is passed through $\pmb{d}$, the resulting output is $\pmb{d}(y') \in (-\infty, \infty)$. To generate a difficulty score $\in (0, 1)$, we squash $L_d^{y'}=-\pmb{d}(y')$ by a sigmoid function~($\sigma$). The difficulty score $\sigma (L_d^{y'})$ is conceptually analogous to the uncertainty $(1-p_t)$ in Focal Loss, so that it is multiplied with the regular loss $L$ to form the Adversarial Focal Loss~(Eq.~\ref{eq_afl}). The intuition behind this formulation can be found in the subsequent section ``Interpretation of the Loss''.
\begin{equation}
    \textrm{AFL}(x) = \textrm{stop}\_\textrm{gradient} (\sigma (L_d^{y'})) L(x)
    \label{eq_afl}
\end{equation}
As shown in Eq.~\ref{eq_afl}, the gradient of the difficulty score is detached, which means the difficulty score is only used as a multiplicative coefficient in the AFL formulation.

Please note that when batch size > 1, the batch-averaging takes place over AFL rather than over $L_d^{y'}$, which ensures that each sample is weighed by its own difficulty score, not by the average difficulty score over that batch. The pseudo-code can be found in Appendix section~\ref{appx_pseudo_code}.

\paragraph{Discriminator}
Since the specific architecture of the discriminator is not the main focus of this paper, we did not spend effort on carefully designing $\pmb{d}$. Instead, we used a light-weight off-the-shelf discriminator from a publicly available implementation~\cite{GAN_Github}.

Following the WGAN training paradigm, we are using gradient penalty~\cite{WGAN-GP} to regularize the discriminator. The discriminator loss is defined as $L_d^{y}-L_d^{y'}+gp\_term$, where $gp\_term$ is the gradient penalty term~\cite{WGAN-GP}. The loss term $L_d^{y}-L_d^{y'}$ can be rewritten as $-\pmb{d}(y)+\pmb{d}(y')$. It guides the training in the same direction as a Cross Entropy objective $\textrm{CE}(p(y), 1) + \textrm{CE}(p(y'), 0)$ does, yet it offers a smoother gradient and higher stability during training~\cite{WGAN}.

\paragraph{Interpretation of the Loss}
Due to the training objective of $\pmb{d}$, it learns to produce a larger $\pmb{d}(y')$ if the representation appears of higher quality (in the sense that it looks like a ground truth representation) and a smaller $\pmb{d}(y')$ otherwise. As a result, a low-quality representation~(i.e., a hard example) would lead to a small $\pmb{d}(y')$, thus a large $-\pmb{d}(y')$, and eventually a difficulty score $\sigma (L_d^{y'})$ closer to 1. On the other hand, an easy example will yield a difficulty score closer to 0.

\paragraph{Topology Extractor}
We used a topology extractor to condense the geometric relations among keypoints within the same object, following the idea from~\cite{KeypointTopology}. It provides a more efficient representation of the keypoint patterns compared to the raw feature maps. Consequently, it reduces the computational cost. For details, see Appendix section~\ref{appx_topo_extractor}.

\section[Empirical Studies]{Empirical Studies\footnote{All experiments were completed on one NVIDIA A100 GPU, expect for two models (R-152 and H-48) for the COCO dataset which required two A100 GPUs.}}

\paragraph{Keypoint detection results on the MPII dataset}

We trained a widely-accepted baseline model~\cite{PoseResNet} using the official implementation~\cite{SimpleBaseline_Github} with AFL integrated, on the MPII dataset~\cite{Dataset_MPII}. The training settings are exactly the same as previously described~\cite{SimpleBaseline}. The standard metric~\cite{Dataset_MPII} is reported in Table~\ref{tb_mpii_val}. Compared to the baseline model, the adoption of AFL provides marginal yet consistent improvements on keypoint detection over all body parts.

\begin{table}[htb!]
  \centering
  \caption{\textbf{Keypoint detection results evaluated on the MPII val dataset.}
    \\Architecture acronyms: R-$x$ = ResNet-$x$-FPN.}
  \begin{tabular}{M{36pt}M{40pt}M{25pt}M{40pt}M{25pt}M{25pt}M{20pt}M{20pt}M{25pt}M{25pt}}
    \toprule
    w/ AFL? & Arch &
    Head~$\uparrow$ & Shoulder~$\uparrow$ & Elbow~$\uparrow$ & Wrist~$\uparrow$ & Hip~$\uparrow$ & Knee~$\uparrow$ & Ankle~$\uparrow$ & Mean~$\uparrow$\\
    \midrule
    \rowcolor{BaselineShade}
    & R-50~\cite{PoseResNet} &
    96.4 & 95.3 & 89.0 & 83.2 & 88.4 & 84.0 & \textbf{79.6} & 88.5\\
    \rowcolor{AFLShade}
    \checkmark & R-50~\cite{PoseResNet} &
    \textbf{96.9} & \textbf{95.5} & \textbf{89.1} & \textbf{83.7} & \textbf{88.9} & \textbf{84.5} & \textbf{79.6} & \textbf{88.9}\\
    \bottomrule
  \end{tabular}
  \label{tb_mpii_val}
\end{table}

\paragraph{Keypoint detection results on the COCO dataset}

We trained 4 models of varying architectures and capacities~\cite{PoseResNet, PoseHRNet} using the official codebase~\cite{HRNet_Github} with AFL integrated on the COCO dataset~\cite{Dataset_COCO}. The training settings are identical to what was previous described~\cite{PoseHRNet}. The standard metric, Average Precision scores, are reported in Table~\ref{tb_coco_val}. This evaluation metric is based on Object Keypoint Similarity~(OKS), a measure of keypoint detection accuracy. Details can be found in~\cite{PoseHRNet}. Specifically, AP$^{50}$/AP$^{75}$ are the average precision at OKS=0.5/OKS=0.75, AP$^M$/AP$^L$ are the average precision for medium/large objects, and AP is the mean average precision at 10 OKS positions evenly spaced between 0.5 and 0.95.

The adoption of AFL increased the mean AP score by 1.5 to 2.0 points across the 4 models. The subcategories AP$^{50}$, AP$^{75}$, and AP$^M$ are also improved by varying amounts.

One intriguing observation is that despite the general improvements, AFL slightly suppresses the average precision on large objects~(AP$^L$). We believe it's because the keypoints on large objects are generally easier to identify, and thus large objects are usually easy samples. As AFL emphasizes hard examples, performance on easy examples may be slightly sacrificed for an overall improvement.

\begin{table}[htb!]
\centering
  \caption{\textbf{Keypoint detection results evaluated on the COCO val dataset.}
  For rows with AFL, the additional computational costs during training are accounted for. We used a light-weight discriminator with 0.0097 M parameters and 0.00056 GFLOPs, and hence the reported Params and FLOPs are not affected.
  Improvements/degradations beyond 1.0 point are highlighted in \color{OliveGreen}{green}\color{black}{/}\color{Bittersweet}{red}\color{black}{.}\\
  $^\dagger$ Input image resized such that the short side is 800 pixels~\cite{MaskRCNN}.\\
  Architecture acronyms: R-$x$ = ResNet-$x$-FPN; X-$x$=ResNeXt-x-FPN; H-$x$ = HRNetV2p-W$x$.}
  \begin{tabular}{M{36pt}M{44pt}M{40pt}M{30pt}M{24pt}M{23pt}M{23pt}M{23pt}M{23pt}M{23pt}}
    \toprule
    w/ AFL? & Arch &
    Input size & Params & FLOPs &
    AP~$\uparrow$ & AP$^{50}$~$\uparrow$ & AP$^{75}$~$\uparrow$ & AP$^{M}$~$\uparrow$ & AP$^{L}$~$\uparrow$\\
    \midrule
    & R-50~\cite{MaskRCNN} &
    $^\dagger$800$\times$800 & - & - & 65.1 & 86.6 & 70.9 & 59.9 & 73.6\\
    & R-50~\cite{Posefix} &
    384$\times$288 & 102 M & - & 72.1 & 88.5 & 78.3 & 68.6 & 78.2\\
    & R-101~\cite{NonLocalNN} &
    224$\times$224 & 51.8 M & - & 66.5 & 87.3 & 72.8 & - & -\\
    & R-101~\cite{MaskRCNN} &
    $^\dagger$800$\times$800 & - & - & 66.1 & 87.7 & 71.7 & 60.5 & 75.0\\
    & X-101~\cite{MaskRCNN} &
    $^\dagger$800$\times$800 & - & - & 70.4 & 89.3 & 76.8 & 65.8 & 78.1\\
    
    \rowcolor{BaselineShade}
    & R-50~\cite{PoseResNet} &
    256$\times$192 & 34.0~M & 9.0~G &
    70.4 & 88.6 & 78.3 & 67.1 & \textbf{77.2}\\
    \rowcolor{AFLShade}
    &&&&&
    \textbf{72.0} & \textbf{92.5} & \textbf{79.3} & \textbf{69.5} & 76.1\\
    \rowcolor{AFLShade}
    \multirow{-2}{*}{\checkmark} &
    \multirow{-2}{*}{R-50~\cite{PoseResNet}} &
    \multirow{-2}{*}{256$\times$192} &
    \multirow{-2}{*}{34.0~M} &
    \multirow{-2}{*}{9.0~G} &
    (\color{OliveGreen}{+1.6}\color{black}{)}&
    (\color{OliveGreen}{+2.9}\color{black}{)}&
    (\color{OliveGreen}{+1.0}\color{black}{)}&
    (\color{OliveGreen}{+2.4}\color{black}{)}&
    (-0.9)\\
    
    \rowcolor{BaselineShade}
    & H-32~\cite{PoseHRNet} &
    256$\times$192 & 28.5~M & 7.1~G &
    74.4 & 90.5 & 81.9 & 70.8 & \textbf{81.0}\\
    \rowcolor{AFLShade}
    &&&&&
    \textbf{76.1} & \textbf{93.6} & \textbf{83.5} & \textbf{73.2} & 80.5\\
    \rowcolor{AFLShade}
    \multirow{-2}{*}{\checkmark} &
    \multirow{-2}{*}{H-32~\cite{PoseHRNet}} &
    \multirow{-2}{*}{256$\times$192} &
    \multirow{-2}{*}{28.5~M} &
    \multirow{-2}{*}{7.1~G} &
    (\color{OliveGreen}{+1.7}\color{black}{)}&
    (\color{OliveGreen}{+3.1}\color{black}{)}&
    (\color{OliveGreen}{+1.6}\color{black}{)}&
    (\color{OliveGreen}{+2.4}\color{black}{)}&
    (-0.5)\\

    \rowcolor{BaselineShade}
    & R-152~\cite{PoseResNet} &
    384$\times$288 & 68.6~M & 35.5~G &
    74.3 & 89.6 & 81.1 & 70.5 & \textbf{81.6}\\
    \rowcolor{AFLShade}
    &&&&&
    \textbf{75.8} & \textbf{92.6} & \textbf{82.5} & \textbf{72.9} & 80.4 \\
    \rowcolor{AFLShade}
    \multirow{-2}{*}{\checkmark} &
    \multirow{-2}{*}{R-152~\cite{PoseResNet}} &
    \multirow{-2}{*}{384$\times$288} &
    \multirow{-2}{*}{68.6~M} &
    \multirow{-2}{*}{35.5~G} &
    (\color{OliveGreen}{+1.5}\color{black}{)}&
    (\color{OliveGreen}{+3.0}\color{black}{)}&
    (\color{OliveGreen}{+1.4}\color{black}{)}&
    (\color{OliveGreen}{+2.4}\color{black}{)}&
    (\color{Bittersweet}{-1.2}\color{black}{)}\\

    \rowcolor{BaselineShade}
    & H-48~\cite{PoseHRNet} &
    384$\times$288 & 63.6~M & 32.9~G &
    76.3 & 90.8 & 82.9 & 72.3 & \textbf{83.4}\\
    \rowcolor{AFLShade}
    &&&&&
    \textbf{78.3} & \textbf{93.6} & \textbf{84.9} & \textbf{75.4} & 83.3\\
    \rowcolor{AFLShade}
    \multirow{-2}{*}{\checkmark} &
    \multirow{-2}{*}{H-48~\cite{PoseHRNet}} &
    \multirow{-2}{*}{384$\times$288} &
    \multirow{-2}{*}{63.6~M} &
    \multirow{-2}{*}{32.9~G} &
    (\color{OliveGreen}{+2.0}\color{black}{)}&
    (\color{OliveGreen}{+2.8}\color{black}{)}&
    (\color{OliveGreen}{+2.0}\color{black}{)}&
    (\color{OliveGreen}{+3.1}\color{black}{)}&
    (-0.1)\\

  \bottomrule
  \end{tabular}
  \label{tb_coco_val}
\end{table}

\paragraph{Keypoint detection results on medical images}

Besides evaluations on standard keypoint detection benchmarks with natural images, we further examined the effectiveness of AFL on medical images.

We trained Feature Pyramid Networks~\cite{FPN} with and without AFL on a restricted X-ray dataset. The dataset defined 36 anatomical landmarks per patient, and was used to train keypoint detection models for diagnostic purposes (diagnostic task is not shown as it is outside the scope of this paper).

The major bottleneck of these keypoint detection models was the false negative rate. The version without AFL falsely generated a considerable number of low-response feature maps. With the adoption of AFL and everything else untouched, the number of false negatives was reduced by approximately a half~(Table.~\ref{tb_internal}).

\begin{table}[htb!]
  \centering
  \caption{\textbf{Keypoint detection results evaluated on a medical dataset.} Models are evaluated on 51 X-ray images with keypoints annotated on 36 anatomical landmarks. These images are pre-processed through 4 different protocols leading to 4 different contrasts that are separately used for evaluation.}
  \begin{tabular}{M{60pt}M{60pt}M{40pt}M{40pt}M{80pt}M{40pt}}
    \toprule
    w/ AFL? & Arch & Contrast & Illustration & False Negative~$\downarrow$ & Total\\
    \midrule
    \rowcolor{BaselineShade}
    & FPN~\cite{FPN} & higher & & 98 \hspace{33pt}~ & 1827\\
    \rowcolor{AFLShade}
    \checkmark & FPN~\cite{FPN} & higher &
    \multirow{-2}{*}{\includegraphics[width=32pt, height=21pt]{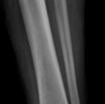}} &
    \textbf{27}~(\color{OliveGreen}{--~72\%}) &
    1827\\
    \rowcolor{BaselineShade}
    & FPN~\cite{FPN} & high & & 85 \hspace{33pt}~ & 1827\\
    \rowcolor{AFLShade}
    \checkmark & FPN~\cite{FPN} & high &
    \multirow{-2}{*}{\includegraphics[width=32pt, height=21pt]{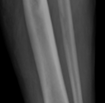}} &
    \textbf{32}~(\color{OliveGreen}{--~62\%}) &
    1827 \\
    \rowcolor{BaselineShade}
    & FPN~\cite{FPN} & low & & 89 \hspace{33pt}~ & 1827\\
    \rowcolor{AFLShade}
    \checkmark & FPN~\cite{FPN} & low &
    \multirow{-2}{*}{\includegraphics[width=32pt, height=21pt]{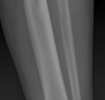}} &
    \textbf{43}~(\color{OliveGreen}{--~52\%}) &
    1827\\
    \rowcolor{BaselineShade}
    & FPN~\cite{FPN} & lower & & 95 \hspace{33pt}~ & 1827\\
    \rowcolor{AFLShade}
    \checkmark & FPN~\cite{FPN} & lower &
    \multirow{-2}{*}{\includegraphics[width=32pt, height=21pt]{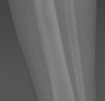}} &
    \textbf{69}~(\color{OliveGreen}{--~27\%}) &
    1827\\
    \bottomrule
  \end{tabular}
  \label{tb_internal}
\end{table}

\paragraph{Exploratory experiment: applying AFL for image classification}

As is evident by its design~(Fig.~\ref{fig_afl_framework_kp}), AFL has potentials beyond keypoint detection tasks. To demonstrate this potential, we evaluated AFL on image classification as a proof-of-concept.

We trained a wide-ResNet~(WRN)~\cite{WideResNet} on CIFAR-100~\cite{CIFAR} using Cross Entropy as the loss function, and keeping Focal Loss and AFL as variables for integration. The results are shown in Table~\ref{tb_cifar100}.

For AFL in image classification, since the input to the discriminator was a one-dimensional probability vector instead of a matrix/image, we used a fully-connected discriminator instead of a convolutional network.

While we were unable to run extensive experiments on bigger datasets (e.g., ImageNet~\cite{ImageNet}) due to limited budget, our exploratory experiments demonstrate AFL's potential on tasks beyond keypoint detection. Focal Loss and AFL both provide noticeable benefits on top of Cross Entropy itself, yet the performance improvement from AFL is marginally bigger.

Interestingly, based on our experiments, using Focal Loss and AFL together does not lead to more favorable results. One possible reason could the over-emphasis of the harder examples while not focusing enough on easy examples.

\begin{table}[htb!]
  \centering
  \caption{\textbf{Image classification results on CIFAR-100.} Each entry is averaged over 3 runs. }
  \begin{tabular}{M{60pt}M{60pt}M{120pt}M{80pt}}
    \toprule
    w/ AFL? & w/ Focal Loss? & Arch & Top-1 Accuracy $\uparrow$\\
    \midrule
     & & WSN-28-10, orig.~\cite{WideResNet} & 79.50 \hspace{27pt}~\\
    \rowcolor{BaselineShade}
     & & WSN-28-10, our impl. & 82.14 \footnotesize~$\pm 0.06$\\
    \rowcolor{AFLShade}
    $\checkmark$ & &  WSN-28-10, our impl. & \textbf{82.86} \footnotesize~$\pm 0.13$\\
    \rowcolor{BaselineShade}
     & $\checkmark$ & WSN-28-10, our impl. & 82.38 \footnotesize~$\pm 0.28$\\
    \rowcolor{AFLShade}
    $\checkmark$ & $\checkmark$ & WSN-28-10, our impl. & 81.83 \footnotesize~$\pm 0.16$\\
    \bottomrule
  \end{tabular}
  \label{tb_cifar100}
\end{table}

\section{Behaviors of the difficulty score}
\label{sec_behavior_diff_score}
As we claimed in section~3, we designed the difficulty score $\sigma (L_d^{y'})$ in our AFL formulation to serve as a multiplicative coefficient to up-weigh the hard examples and down-weigh the easy examples during training. We wanted to investigate whether it achieved this purpose. To that end, we examined the history of the difficulty score on easy versus hard examples during training.

Based on our speculation, we would anticipate the following trend:
\begin{enumerate}
    \item At the very beginning of training, due to limited discriminative power, $\pmb{d}$ will produce similar difficulty scores for all data samples. But the scores may cluster at any arbitrary value~$\in (0, 1)$ depending on the initialization of $\pmb{d}$.
    \item  As training progresses, $\pmb{f}$ begins to generate higher-quality keypoint detections on easy examples before they can handle hard examples. At this stage, difficulty scores will start to diverge -- the score will be lower for easy examples and higher for hard examples.
\end{enumerate}

\begin{figure}[tb!]
  \centering
  \includegraphics[width=370pt]{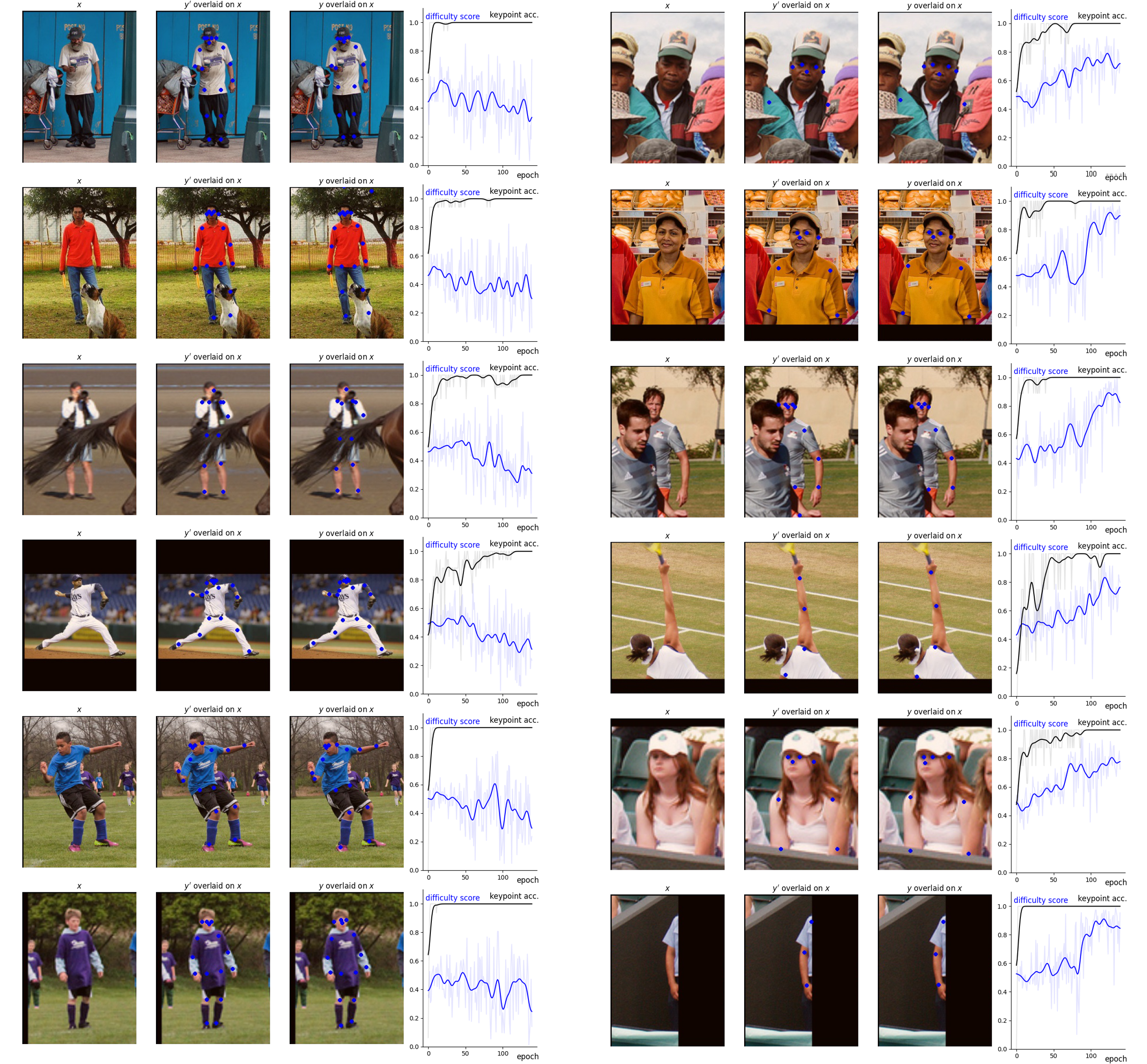}
  \caption{\textbf{Progression of the difficulty score $\sigma (L_d^{y'})$ for easy versus hard examples.} Left: easy examples; Right: hard examples. Within left/right, the 4 columns respectively represent input image, keypoint prediction, keypoint ground truth, and the difficulty score over epoch. Curves are Gaussian-smoothed for cleaner display, while the raw curves are shown in background with higher transparency. \textit{It can be seen that a ``hard example'' does not necessarily imply bad performance.}}
  \label{fig_difficulty_score_main}
\end{figure}

During the COCO training, we plotted the progression of the difficulty scores over time~(Fig.~\ref{fig_difficulty_score_main}). We randomly selected a few data samples to track their difficulty scores, and then grouped them into easy and hard examples based on the discriminator's response~(Fig.~\ref{fig_difficulty_score_main}, left and right). We find that $\pmb{d}$ is able to judge the difficulty of data samples in a sensible manner. In general, easy examples~(Fig.~\ref{fig_difficulty_score_main}, left) exhibit well-exposed people in normal stances with keypoints visible over the full body. Hard examples~(Fig.~\ref{fig_difficulty_score_main}, right), on the other hand, are people that manifest unusual postures or have their body parts significantly occluded, which lead to abnormal keypoint relations or many missing keypoints.

The behaviors of the difficulty scores shown in our experiments are well aligned with our expectations, providing a glimpse into the underlying mechanism of the AFL method.

\section{Conclusion and Future Works}
In this paper, we first described the need for a Focal Loss equivalent in keypoint detection, and we explained why Focal Loss cannot be trivially adapted outside the classification domain. Then, we presented a novel approach called Adversarial Focal Loss~(AFL) that is semantically equivalent to Focal Loss in non-classification tasks. We showed AFL's empirical effectiveness on three keypoint detection datasets and one image classification dataset. Finally, we investigated the behavior of AFL during training to verify its mechanism.

As for future work, we have identified two potential areas for further investigation.

\paragraph{Stronger regularization on the range of $L_d^{y'}$}
The AFL method relies on the assignment of a proper difficulty score $\sigma(L_d^{y'})$ to each input sample. This consequently requires $L_d^{y'}$ to fall in a reasonable range. The current WGAN-like objective $L_d^{y}-L_d^{y'}+gp\_term$ does not pose a strict enforcement on this. For example, $\{L_d^{y}=-50, L_d^{y'}-60\}$ and $\{L_d^{y}=10, L_d^{y'}=0\}$ both result in the same $L_d^{y}-L_d^{y'}$, but the former is more likely to produce difficulty scores close to 1 for all input samples, while the latter is more likely to produce diverse difficulty scores between 0 and 1. Admittedly, the gradient penalty term help regularize $L_d^{y}-L_d^{y'}$ towards zero-mean, but it is relatively weak and indirect. We believe a stronger regularization on the range of $L_d^{y'}$ may be preferable.

\paragraph{Extending AFL to other tasks}

It is easy to notice that AFL can be applied to a wider domain than Focal Loss thanks to its design~(Fig.~\ref{fig_afl_framework_kp}). Besides tasks that we have already studied, any task that follows the formulation in the upper section of Fig.~\ref{fig_afl_framework_kp} may find AFL beneficial. This would include denoising, semantic segmentation, superresolution, domain adaptation, among many others. A particularly interesting possibility is AFL in GANs, i.e., using AFL to dynamically adjust per-sample generator loss in tasks such as image generation, style transfer, etc. While this may lead to complications in optimization, the formulation of AFL shows no objection to such challenges.

\newpage
\bibliography{main}

\begin{thebibliography}{10}

\bibitem{ClassImbalanceExp1}
Jianxiong Xiao, James Hays, Krista~A Ehinger, Aude Oliva, and Antonio Torralba.
\newblock Sun database: Large-scale scene recognition from abbey to zoo.
\newblock In {\em 2010 IEEE computer society conference on computer vision and
  pattern recognition}, pages 3485--3492. IEEE, 2010.

\bibitem{ClassImbalanceExp2}
Brian Mac~Namee, Padraig Cunningham, Stephen Byrne, and Owen~I Corrigan.
\newblock The problem of bias in training data in regression problems in
  medical decision support.
\newblock {\em Artificial intelligence in medicine}, 24(1):51--70, 2002.

\bibitem{ClassImbalanceExp3}
Claire Cardie and Nicholas Howe.
\newblock Improving minority class prediction using case-specific feature
  weights.
\newblock In {\em International Conference on Machine Learning}, 1997.

\bibitem{ClassImbalanceDataRand1}
Mateusz Buda, Atsuto Maki, and Maciej~A Mazurowski.
\newblock A systematic study of the class imbalance problem in convolutional
  neural networks.
\newblock {\em Neural Networks}, 106:249--259, 2018.

\bibitem{ClassImbalanceDataRand2}
Jason Van~Hulse, Taghi~M Khoshgoftaar, and Amri Napolitano.
\newblock Experimental perspectives on learning from imbalanced data.
\newblock In {\em Proceedings of the 24th international conference on Machine
  learning}, pages 935--942, 2007.

\bibitem{ClassImbalanceDataSoph1}
Inderjeet Mani and I~Zhang.
\newblock knn approach to unbalanced data distributions: a case study involving
  information extraction.
\newblock In {\em Proceedings of workshop on learning from imbalanced
  datasets}, volume 126, pages 1--7. ICML, 2003.

\bibitem{ClassImbalanceDataSoph2}
Miroslav Kubat, Stan Matwin, et~al.
\newblock Addressing the curse of imbalanced training sets: one-sided
  selection.
\newblock In {\em Icml}, volume~97, page 179. Citeseer, 1997.

\bibitem{ClassImbalanceAlg1}
Bartosz Krawczyk.
\newblock Learning from imbalanced data: open challenges and future directions.
\newblock {\em Progress in Artificial Intelligence}, 5(4):221--232, 2016.

\bibitem{ClassImbalanceAlg2}
Charles~X Ling and Victor~S Sheng.
\newblock Cost-sensitive learning and the class imbalance problem.
\newblock {\em Encyclopedia of machine learning}, 2011:231--235, 2008.

\bibitem{FocalLoss}
Tsung-Yi Lin, Priya Goyal, Ross Girshick, Kaiming He, and Piotr Doll{\'a}r.
\newblock Focal loss for dense object detection.
\newblock In {\em Proceedings of the IEEE international conference on computer
  vision}, pages 2980--2988, 2017.

\bibitem{FocalLossApp1}
Giang~Son Tran, Thi~Phuong Nghiem, Van~Thi Nguyen, Chi~Mai Luong, and
  Jean-Christophe Burie.
\newblock Improving accuracy of lung nodule classification using deep learning
  with focal loss.
\newblock {\em Journal of healthcare engineering}, 2019, 2019.

\bibitem{FocalLossApp2}
Jishnu Mukhoti, Viveka Kulharia, Amartya Sanyal, Stuart Golodetz, Philip Torr,
  and Puneet Dokania.
\newblock Calibrating deep neural networks using focal loss.
\newblock {\em Advances in Neural Information Processing Systems},
  33:15288--15299, 2020.

\bibitem{FocalLossApp3}
Hei Law and Jia Deng.
\newblock Cornernet: Detecting objects as paired keypoints.
\newblock In {\em Proceedings of the European conference on computer vision
  (ECCV)}, pages 734--750, 2018.

\bibitem{FocalLossApp4}
Zhi Tian, Chunhua Shen, Hao Chen, and Tong He.
\newblock Fcos: Fully convolutional one-stage object detection.
\newblock In {\em Proceedings of the IEEE/CVF international conference on
  computer vision}, pages 9627--9636, 2019.

\bibitem{FocalLossApp5}
Mingsheng Long, Zhangjie Cao, Jianmin Wang, and Michael~I Jordan.
\newblock Conditional adversarial domain adaptation.
\newblock {\em Advances in neural information processing systems}, 31, 2018.

\bibitem{FocalLossApp6}
Xingyi Zhou, Dequan Wang, and Philipp Kr{\"a}henb{\"u}hl.
\newblock Objects as points.
\newblock {\em arXiv preprint arXiv:1904.07850}, 2019.

\bibitem{KeypointFace1}
Xiangxin Zhu and Deva Ramanan.
\newblock Face detection, pose estimation, and landmark localization in the
  wild.
\newblock In {\em 2012 IEEE conference on computer vision and pattern
  recognition}, pages 2879--2886. IEEE, 2012.

\bibitem{KeypointFace2}
Yi~Sun, Xiaogang Wang, and Xiaoou Tang.
\newblock Deep convolutional network cascade for facial point detection.
\newblock In {\em Proceedings of the IEEE conference on computer vision and
  pattern recognition}, pages 3476--3483, 2013.

\bibitem{KeypointPose1}
Zigang Geng, Ke~Sun, Bin Xiao, Zhaoxiang Zhang, and Jingdong Wang.
\newblock Bottom-up human pose estimation via disentangled keypoint regression.
\newblock In {\em Proceedings of the IEEE/CVF Conference on Computer Vision and
  Pattern Recognition}, pages 14676--14686, 2021.

\bibitem{KeypointPose2}
Adrian Bulat and Georgios Tzimiropoulos.
\newblock Human pose estimation via convolutional part heatmap regression.
\newblock In {\em European Conference on Computer Vision}, pages 717--732.
  Springer, 2016.

\bibitem{KeypointPose3}
Xiao Chu, Wei Yang, Wanli Ouyang, Cheng Ma, Alan~L Yuille, and Xiaogang Wang.
\newblock Multi-context attention for human pose estimation.
\newblock In {\em Proceedings of the IEEE conference on computer vision and
  pattern recognition}, pages 1831--1840, 2017.

\bibitem{KeypointMedical1}
Jun Zhang, Mingxia Liu, and Dinggang Shen.
\newblock Detecting anatomical landmarks from limited medical imaging data
  using two-stage task-oriented deep neural networks.
\newblock {\em IEEE Transactions on Image Processing}, 26(10):4753--4764, 2017.

\bibitem{KeypointMedical2}
Qingsong Yao, Quan Quan, Li~Xiao, and S~Kevin~Zhou.
\newblock One-shot medical landmark detection.
\newblock In {\em International Conference on Medical Image Computing and
  Computer-Assisted Intervention}, pages 177--188. Springer, 2021.

\bibitem{KeypointImbalance1}
Kemal Oksuz, Baris~Can Cam, Sinan Kalkan, and Emre Akbas.
\newblock Imbalance problems in object detection: A review.
\newblock {\em IEEE transactions on pattern analysis and machine intelligence},
  43(10):3388--3415, 2020.

\bibitem{KeypointImbalance2}
Yu~Li, Tao Wang, Bingyi Kang, Sheng Tang, Chunfeng Wang, Jintao Li, and Jiashi
  Feng.
\newblock Overcoming classifier imbalance for long-tail object detection with
  balanced group softmax.
\newblock In {\em Proceedings of the IEEE/CVF conference on computer vision and
  pattern recognition}, pages 10991--11000, 2020.

\bibitem{KeypointImbalance3}
Yuan Yao, Yasamin Jafarian, and Hyun~Soo Park.
\newblock Monet: Multiview semi-supervised keypoint detection via epipolar
  divergence.
\newblock In {\em Proceedings of the IEEE/CVF International Conference on
  Computer Vision}, pages 753--762, 2019.

\bibitem{SIFT}
David~G Lowe.
\newblock Distinctive image features from scale-invariant keypoints.
\newblock {\em International journal of computer vision}, 60(2):91--110, 2004.

\bibitem{SURF}
Herbert Bay, Tinne Tuytelaars, and Luc~Van Gool.
\newblock Surf: Speeded up robust features.
\newblock In {\em European conference on computer vision}, pages 404--417.
  Springer, 2006.

\bibitem{ORB}
Ethan Rublee, Vincent Rabaud, Kurt Konolige, and Gary Bradski.
\newblock Orb: An efficient alternative to sift or surf.
\newblock In {\em 2011 International conference on computer vision}, pages
  2564--2571. Ieee, 2011.

\bibitem{Hourglass}
Alejandro Newell, Kaiyu Yang, and Jia Deng.
\newblock Stacked hourglass networks for human pose estimation.
\newblock In {\em European conference on computer vision}, pages 483--499.
  Springer, 2016.

\bibitem{FPN}
Tsung-Yi Lin, Piotr Doll{\'a}r, Ross Girshick, Kaiming He, Bharath Hariharan,
  and Serge Belongie.
\newblock Feature pyramid networks for object detection.
\newblock In {\em Proceedings of the IEEE conference on computer vision and
  pattern recognition}, pages 2117--2125, 2017.

\bibitem{MaskRCNN}
Kaiming He, Georgia Gkioxari, Piotr Doll{\'a}r, and Ross Girshick.
\newblock Mask r-cnn.
\newblock In {\em Proceedings of the IEEE international conference on computer
  vision}, pages 2961--2969, 2017.

\bibitem{CPN}
Yilun Chen, Zhicheng Wang, Yuxiang Peng, Zhiqiang Zhang, Gang Yu, and Jian Sun.
\newblock Cascaded pyramid network for multi-person pose estimation.
\newblock In {\em Proceedings of the IEEE conference on computer vision and
  pattern recognition}, pages 7103--7112, 2018.

\bibitem{OpenPose}
Zhe Cao, Tomas Simon, Shih-En Wei, and Yaser Sheikh.
\newblock Realtime multi-person 2d pose estimation using part affinity fields.
\newblock In {\em Proceedings of the IEEE conference on computer vision and
  pattern recognition}, pages 7291--7299, 2017.

\bibitem{AssociativeEmbedding}
Alejandro Newell, Zhiao Huang, and Jia Deng.
\newblock Associative embedding: End-to-end learning for joint detection and
  grouping.
\newblock {\em Advances in neural information processing systems}, 30, 2017.

\bibitem{PersonLab}
George Papandreou, Tyler Zhu, Liang-Chieh Chen, Spyros Gidaris, Jonathan
  Tompson, and Kevin Murphy.
\newblock Personlab: Person pose estimation and instance segmentation with a
  bottom-up, part-based, geometric embedding model.
\newblock In {\em Proceedings of the European conference on computer vision
  (ECCV)}, pages 269--286, 2018.

\bibitem{MultiPoseNet}
Muhammed Kocabas, Salih Karagoz, and Emre Akbas.
\newblock Multiposenet: Fast multi-person pose estimation using pose residual
  network.
\newblock In {\em Proceedings of the European conference on computer vision
  (ECCV)}, pages 417--433, 2018.

\bibitem{PoseResNet}
Bin Xiao, Haiping Wu, and Yichen Wei.
\newblock Simple baselines for human pose estimation and tracking.
\newblock In {\em European Conference on Computer Vision (ECCV)}, 2018.

\bibitem{PoseHRNet}
Jingdong Wang, Ke~Sun, Tianheng Cheng, Borui Jiang, Chaorui Deng, Yang Zhao,
  Dong Liu, Yadong Mu, Mingkui Tan, Xinggang Wang, et~al.
\newblock Deep high-resolution representation learning for visual recognition.
\newblock {\em IEEE transactions on pattern analysis and machine intelligence},
  43(10):3349--3364, 2020.

\bibitem{KeypointRepreOther1}
Aiden Nibali, Zhen He, Stuart Morgan, and Luke Prendergast.
\newblock Numerical coordinate regression with convolutional neural networks.
\newblock {\em arXiv preprint arXiv:1801.07372}, 2018.

\bibitem{KeypointRepreOther2}
Zhengxiong Luo, Zhicheng Wang, Yan Huang, Liang Wang, Tieniu Tan, and Erjin
  Zhou.
\newblock Rethinking the heatmap regression for bottom-up human pose
  estimation.
\newblock In {\em Proceedings of the IEEE/CVF Conference on Computer Vision and
  Pattern Recognition}, pages 13264--13273, 2021.

\bibitem{GAN}
Ian Goodfellow, Jean Pouget-Abadie, Mehdi Mirza, Bing Xu, David Warde-Farley,
  Sherjil Ozair, Aaron Courville, and Yoshua Bengio.
\newblock Generative adversarial nets.
\newblock {\em Advances in neural information processing systems}, 27, 2014.

\bibitem{WGAN}
Martin Arjovsky, Soumith Chintala, and L{\'e}on Bottou.
\newblock Wasserstein generative adversarial networks.
\newblock In {\em International conference on machine learning}, pages
  214--223. PMLR, 2017.

\bibitem{GAN_Github}
E.~Linder-Norén.
\newblock Pytorch implementations of generative adversarial networks.
\newblock \url{https://github.com/eriklindernoren/PyTorch-GAN}, 2018.

\bibitem{WGAN-GP}
Ishaan Gulrajani, Faruk Ahmed, Martin Arjovsky, Vincent Dumoulin, and Aaron~C
  Courville.
\newblock Improved training of wasserstein gans.
\newblock {\em Advances in neural information processing systems}, 30, 2017.

\bibitem{KeypointTopology}
Shaobo Zhang, Wanqing Zhao, Ziyu Guan, Xianlin Peng, and Jinye Peng.
\newblock Keypoint-graph-driven learning framework for object pose estimation.
\newblock In {\em Proceedings of the IEEE/CVF Conference on Computer Vision and
  Pattern Recognition}, pages 1065--1073, 2021.

\bibitem{SimpleBaseline_Github}
X.~Bin.
\newblock Simple baselines for human pose estimation and tracking.
\newblock \url{https://github.com/Microsoft/human-pose-estimation.pytorch},
  2018.

\bibitem{Dataset_MPII}
Mykhaylo Andriluka, Leonid Pishchulin, Peter Gehler, and Bernt Schiele.
\newblock 2d human pose estimation: New benchmark and state of the art
  analysis.
\newblock In {\em Proceedings of the IEEE Conference on computer Vision and
  Pattern Recognition}, pages 3686--3693, 2014.

\bibitem{SimpleBaseline}
Bin Xiao, Haiping Wu, and Yichen Wei.
\newblock Simple baselines for human pose estimation and tracking.
\newblock In {\em Proceedings of the European conference on computer vision
  (ECCV)}, pages 466--481, 2018.

\bibitem{HRNet_Github}
X.~Bin.
\newblock Deep high-resolution representation learning for human pose
  estimation (cvpr 2019).
\newblock \url{https://github.com/leoxiaobin/deep-high-resolution-net.pytorch},
  2019.

\bibitem{Dataset_COCO}
Tsung-Yi Lin, Michael Maire, Serge Belongie, James Hays, Pietro Perona, Deva
  Ramanan, Piotr Doll{\'a}r, and C~Lawrence Zitnick.
\newblock Microsoft coco: Common objects in context.
\newblock In {\em European conference on computer vision}, pages 740--755.
  Springer, 2014.

\bibitem{Posefix}
Gyeongsik Moon, Ju~Yong Chang, and Kyoung~Mu Lee.
\newblock Posefix: Model-agnostic general human pose refinement network.
\newblock In {\em Proceedings of the IEEE/CVF Conference on Computer Vision and
  Pattern Recognition}, pages 7773--7781, 2019.

\bibitem{NonLocalNN}
Xiaolong Wang, Ross Girshick, Abhinav Gupta, and Kaiming He.
\newblock Non-local neural networks.
\newblock In {\em Proceedings of the IEEE conference on computer vision and
  pattern recognition}, pages 7794--7803, 2018.

\bibitem{WideResNet}
Sergey Zagoruyko and Nikos Komodakis.
\newblock Wide residual networks.
\newblock {\em arXiv preprint arXiv:1605.07146}, 2016.

\bibitem{CIFAR}
Alex Krizhevsky, Geoffrey Hinton, et~al.
\newblock Learning multiple layers of features from tiny images.
\newblock {\em technical
  report~\url{https://www.cs.toronto.edu/~kriz/learning-features-2009-TR.pdf}},
  2009.

\bibitem{ImageNet}
Jia Deng, Wei Dong, Richard Socher, Li-Jia Li, Kai Li, and Li~Fei-Fei.
\newblock Imagenet: A large-scale hierarchical image database.
\newblock In {\em 2009 IEEE conference on computer vision and pattern
  recognition}, pages 248--255. Ieee, 2009.

\end{thebibliography}
\bibliographystyle{unsrt}
\newpage
\appendix

\section{Appendix}
\subsection{Pseudo-code of the vanilla method for keypoint detection versus our proposed AFL}
\label{appx_pseudo_code}


\begin{algorithm}
\caption{Vanilla Method}
\SetKwInOut{Req}{Required}
\Req{
\\input images $X$\\
ground truth feature maps $Y$\\
main model $\pmb{f}$\\
$ $\\
}
\For{$(x, y) \in (X, Y)$}{
    $y' \gets \pmb{f}(x)$\\
    $L \gets \textrm{loss\_function}(y, y')$\\
    $\pmb{f} \textrm{ updates w.r.t. } L$
}
\end{algorithm}

\begin{algorithm}
\caption{Proposed AFL Method}
\SetKwInOut{Req}{Required}
\SetKwInOut{Opt}{Optional}
\Req{
\\input images $X$\\
ground truth feature maps $Y$\\
main model $\pmb{f}$\\
discriminator $\pmb{d}$\\
}
\Opt{
\\topology extractor $\pmb{t}$ (required for keypoint detection)\\
$ $\\
}
\For{$(x, y) \in (X, Y)$}{
    $y' \gets \pmb{f}(x)$\\
    $L \gets \textrm{loss\_function}(y, y')$\\
    $ $\\
    $/* \textrm{ The following 2 lines are only relevant to keypoint detection. Skip them otherwise. } */$\\
    $y \gets \pmb{t}(y)$\\
    $y' \gets \pmb{t}(y')$\\
    $ $\\
    $(L_d^y, L_d^{y'}, \textrm{gp\_term}) \gets \textrm{WGAN\_GP}(\pmb{d}, y, y')$\\
    $L_d \gets L_d^y - L_d^{y'} + \textrm{gp\_term}$\\
    $\textrm{AFL} \gets \textrm{stop\_gradient}(\sigma(L_d^{y'})) \cdot L$\\
    $ $\\
    $\pmb{d} \textrm{ updates w.r.t. } L_d$\\
    $\pmb{f} \textrm{ updates w.r.t. } \textrm{AFL}$
}
$ $\\
$/* \textrm{ Detail of WGAN\_GP \cite{WGAN-GP} }(\textrm{Using the default setting: }\lambda = 10) */$\\
$ $\\
\begin{algorithmic}[1]
\Procedure{\textrm{WGAN\_GP}}{$\pmb{d}, y, y'$}
  \State $\alpha \sim U[0, 1]$
  \State $y_{\textrm{mixup}} \gets \alpha \cdot y + (1-\alpha) \cdot y'$
  \State $L_d^y \gets -\pmb{d}(y)$
  \State $L_d^{y'} \gets -\pmb{d}(y')$
  \State $\textrm{gp\_term} \gets \lambda \cdot (||\nabla_{y_{\textrm{mixup}}} \pmb{d}(y_{\textrm{mixup}})||_2-1)^2$\\
$\textrm{\textbf{return} } L_d^y, L_d^{y'}, \textrm{gp\_term}$
\EndProcedure
\end{algorithmic}

\end{algorithm}

\subsection{Topology extractor}
\label{appx_topo_extractor}
The topology extractor is a mapping from the feature space~$\mathbb{R}^{W\times H\times K}$ (for $K$ keypoints) to a set of $N$ adjacency matrices~$\mathbb{R}^{K\times K\times N}$. In our current formulation we have $N=2$, where the first adjacency matrix represents the planar affinity of any two keypoints and the second matrix represents their angular affinity.

In practice, we first instantiate two matrices $M_p$ and $M_a$, both of dimensionality $\mathbb{R}^{K\times K}$, to respectively represent the planar and angular affinities. We then find the coordinates of the centroids of these keypoints $(x, y)_i \textrm{ for } i \in \mathbb{K} \textrm{, where } \mathbb{K}:= [1, 2, ..., K]$. Any keypoint $j$ that does not exist on the feature map and thus does not have a centroid will be ignored, such that $M(j, i)$ and $M(i, j)$ will remain 0 for all $i \in \mathbb{K}$. For notational simplicity, we define $\mathbb{K}^+$ as the subset of $\mathbb{K}$ where keypoints exist.

For planar affinity, we define
\begin{equation}
    M_p(i,j)=
    1 - \frac{||(x, y)_i - (x, y)_j||_2}
    {||(W, H)-(0, 0)||_2}
\end{equation}

For angular affinity, we define
\begin{equation}
    M_a(i, j)=
    \frac{1}{2}+
    \frac{1}{2}\cos{\angle{A(i, j)}}
\end{equation}
To find $\angle{A(i, j)}$, we first compute the coordinates of the ``global centroid'' among all existing keypoints, $\displaystyle(x, y)=\left(\frac{1}{|\mathbb{K}^+|}\sum_{k \in \mathbb{K}^+}x_k, \frac{1}{|\mathbb{K}^+|}\sum_{k \in \mathbb{K}^+}y_k\right)$. Then, we define $\angle{A(i, j)}$ for each $i, j \in \mathbb{K}^+$ as an angle whose vertex is the ``global centroid'' $(x, y)$ and whose two sides pass through $(x, y)_i$ and $(x, y)_j$, respectively.

\newpage
\subsection{Additional cases for section~\ref{sec_behavior_diff_score}}

\vspace{10pt}
\begin{figure}[h!]
  \centering
  \includegraphics[width=370pt]{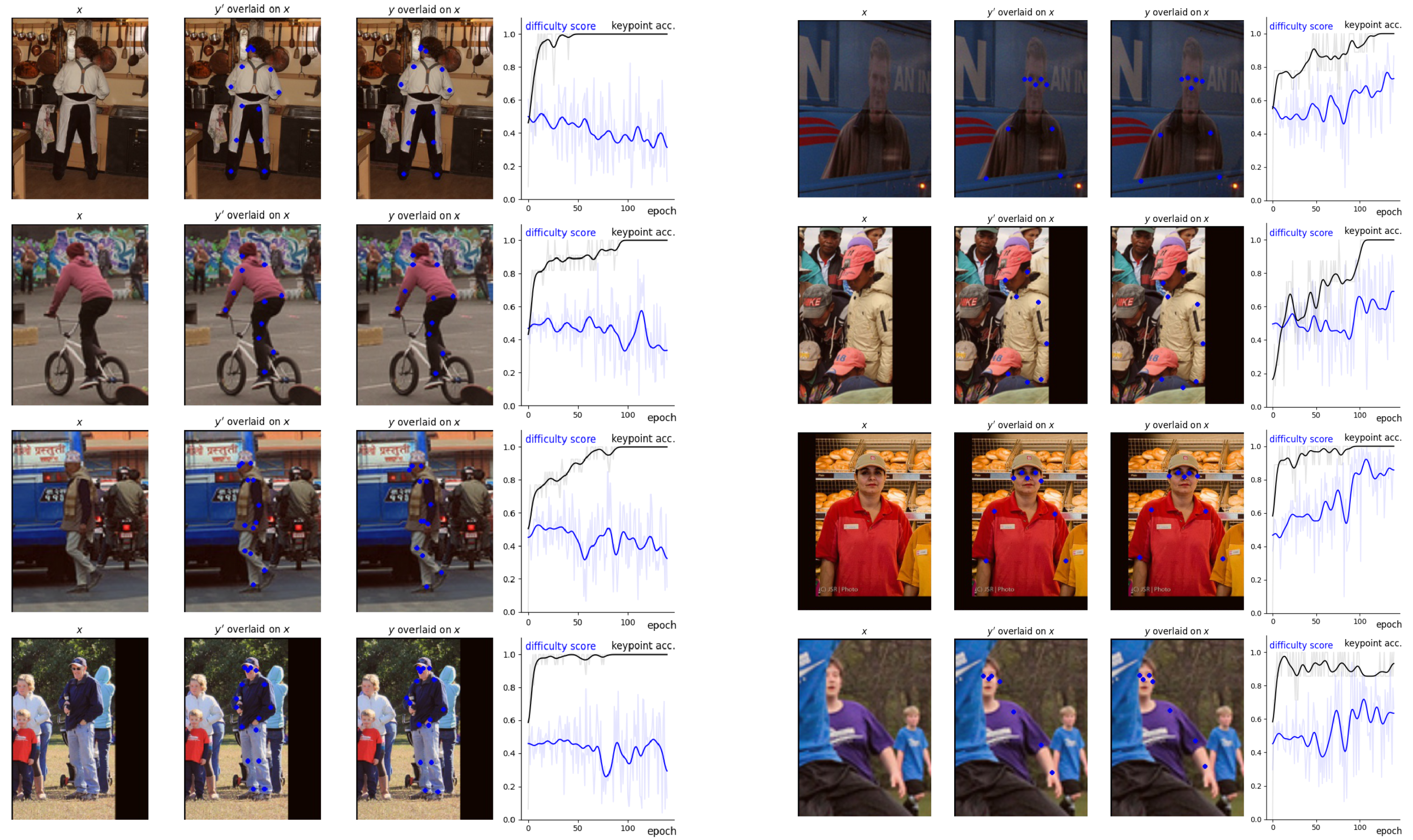}
  \caption{\textbf{Progression of the difficulty score $\sigma (L_d^{y'})$ for easy versus hard examples: additional cases.} Left: easy examples; Right: hard examples. Within left/right, the 4 columns respectively represent input image, keypoint prediction, keypoint ground truth, and the difficulty score over epoch. Curves are Gaussian-smoothed for cleaner display, while the raw curves are shown in background with higher transparency. \textit{Again, it can be seen that a ``hard example'' does not necessarily imply bad performance.}}
\end{figure}

\newpage
\subsection{Failure cases for section~\ref{sec_behavior_diff_score}}

\begin{figure}[h!]
  \centering
  \includegraphics[width=370pt]{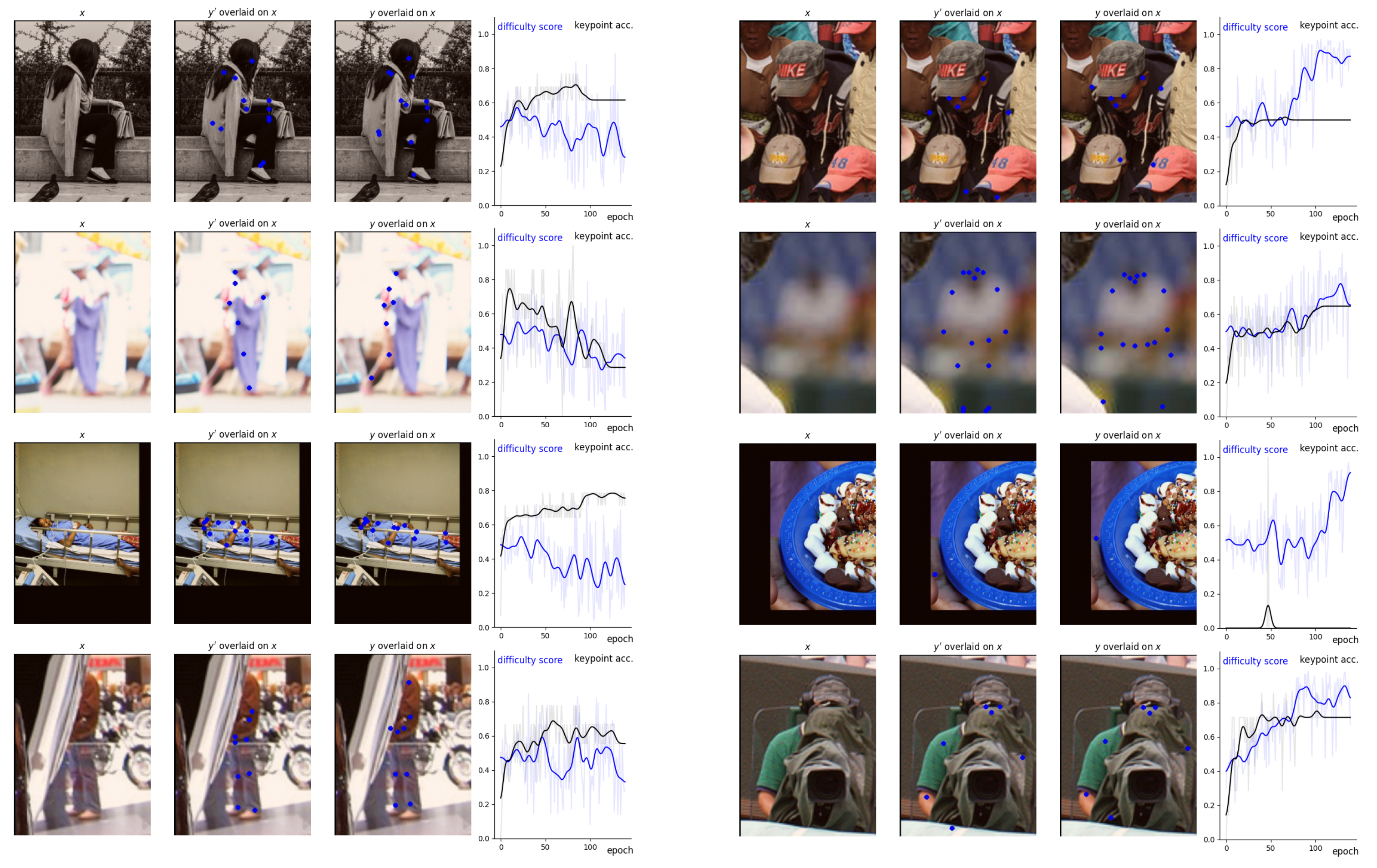}
  \caption{\textbf{Progression of the difficulty score $\sigma (L_d^{y'})$ for easy versus hard examples: failure cases. Examples with low keypoint detection accuracy are selected for this figure.} Left: easy examples; Right: hard examples. Within left/right, the 4 columns respectively represent input image, keypoint prediction, keypoint ground truth, and the difficulty score over epoch. Curves are Gaussian-smoothed for cleaner display, while the raw curves are shown in background with higher transparency. Just like a ``hard example'' does not necessarily imply bad performance, \textit{an ``easy example'' does not necessarily imply good performance.}}
\end{figure}




    

\end{document}